%% file: main.tex
\pdfoutput=1
\documentclass[10pt, logo, twocolumn, copyright, nonumbering]{nvidiatechreport}
\usepackage{caption}
\usepackage{mdframed}
\usepackage[utf8]{inputenc} 
\usepackage[T1]{fontenc}    
\usepackage{hyperref}       
\usepackage{url}            

\usepackage{booktabs}       
\usepackage{amsfonts}       
\usepackage{nicefrac}       
\usepackage{microtype}      
\usepackage[dvipsnames]{xcolor}         
\usepackage{multirow}
\usepackage{multicol}
\usepackage{graphicx}
\usepackage{subcaption}
\usepackage[numbers]{natbib}
\usepackage{tabto}
\usepackage{xspace}
\usepackage{amsmath}
\usepackage{adjustbox}
\usepackage{enumitem}
\usepackage{wrapfig}
\usepackage{dblfloatfix}
\usepackage{algorithm}
\usepackage{algpseudocode}
\usepackage[most]{tcolorbox}   
\usepackage{graphicx}          
\usepackage{hyperref}          
\usepackage{array}             
\usepackage{geometry}          
\usepackage{cuted}  

\newcolumntype{g}{>{\color{gray}}c} 

\title{Minitron-SSM: Efficient Hybrid Language Model Compression through Group-Aware SSM Pruning}
\author{
Ali Taghibakhshi*,
Sharath Turuvekere Sreenivas*,
Saurav Muralidharan*,
Marcin Chochowski*,
Yashaswi Karnati*,
Raviraj Joshi,
Ameya Sunil Mahabaleshwarkar,
Zijia Chen,
Yoshi Suhara,
Oluwatobi Olabiyi,
Daniel Korzekwa,
Mostofa Patwary,
Mohammad Shoeybi,
Jan Kautz,
Bryan Catanzaro,
Ashwath Aithal,
Nima Tajbakhsh,
Pavlo Molchanov
}

\correspondingauthor{X}

\begin{abstract}
\textbf{Abstract:}
Hybrid LLM architectures that combine Attention and State Space Models (SSMs) achieve state-of-the-art accuracy and runtime performance.
Recent work has demonstrated that applying compression and distillation to Attention-only models yields smaller, more accurate models at a fraction of the training cost. In this work, we explore the effectiveness of compressing Hybrid architectures. We introduce a novel group-aware pruning strategy that preserves the structural integrity of SSM blocks and their sequence modeling capabilities. Furthermore, we demonstrate the necessity of such SSM pruning to achieve improved accuracy and inference speed compared to traditional approaches. Our compression recipe combines SSM, FFN, embedding dimension, and layer pruning, followed by knowledge distillation-based retraining, similar to the MINITRON 
technique. Using this approach, we compress the Nemotron-H 8B Hybrid model down to 4B parameters with up to 40x fewer training tokens. The resulting model surpasses the accuracy of similarly-sized models while achieving $\sim$2x faster inference, significantly advancing the Pareto frontier.

\end{abstract}

\begin{document}
\maketitle

\begin{strip}
\begin{center}
\begin{tcolorbox}[colback=gray!3, colframe=gray!30, arc=3pt, boxsep=3pt,
  left=10pt, right=10pt, top=5pt, bottom=5pt, width=0.85\textwidth]
\centering
\textbf{Models on Hugging Face} \\[5pt]
\resizebox{0.85\textwidth}{!}{%
\begin{tabular}{@{}cc@{}}
\raisebox{-0.4\height}{\includegraphics[height=1.1em]{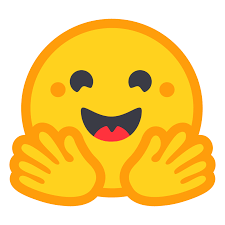}}~%
\href{https://huggingface.co/nvidia/Nemotron-H-4B-Base-8K}{\texttt{Nemotron-H-4B-Base-8K}} &
\raisebox{-0.4\height}{\includegraphics[height=1.1em]{fig/hf_logo.png}}~%
\href{https://huggingface.co/nvidia/Nemotron-H-4B-Instruct-128K}{\texttt{Nemotron-H-4B-Instruct-128K}} \\
\end{tabular}%
}
\end{tcolorbox}
\end{center}
\end{strip}

\input{tex/introduction.tex}
\input{tex/background.tex}
\input{tex/method.tex}
\input{tex/results.tex}

\section*{Conclusions}

In this paper, we present Nemotron-H 4B, a compressed hybrid language model that combines Attention and State Space Models (SSMs) to achieve state-of-the-art accuracy and efficiency. By leveraging a novel group-aware pruning strategy for Mamba layers combined with structured pruning of FFN neurons and embedding dimensions, and knowledge distillation, we reduce the model size by 50\% while retaining over 96\% of the original 8B model’s accuracy, with up to 40× fewer training tokens.

Nemotron-H 4B advances the accuracy-efficiency Pareto frontier, achieving $\sim$2× faster inference and 2.6\% higher accuracy across a diverse set of tasks. The instruction-tuned variant further excels in long-context reasoning (up to 128K tokens) and tool-use applications, making it a compelling choice for resource-constrained deployments. By open-sourcing our compression
recipe, we provide a practical blueprint for efficient hybrid model development.


\section{Acknowledgments}
This work would not have been possible without contributions from many people at NVIDIA. To mention a few:

Akhiad Bercovich, Brandon Norick, Boris Ginsburg, Chengyu Dong, Dan Su, Deepak Narayanan, Dima Rekesh, Duncan Riach, Eileen Long, Elad Segal, Eric Harper, Izik Golan, Jared Casper, John Kamalu, Joseph Jennings, Jupinder Parmar, Kezhi Kong, Markus Klieg, Ran El-Yaniv, Roger Waleffe, Sanjeev Satheesh, Shrimai Prabhumoye, Syeda Nahida Akter, Tomer Ronen, Ying Lin.

{
  \small
  \bibliographystyle{plainnat}
  \bibliography{paper}
}

\end{document}

%% file: tex/introduction.tex
\begin{figure*}[h!]
    \centering
    \begin{subfigure}[t]{0.4775\textwidth}
        \centering
        \includegraphics[width=\linewidth]{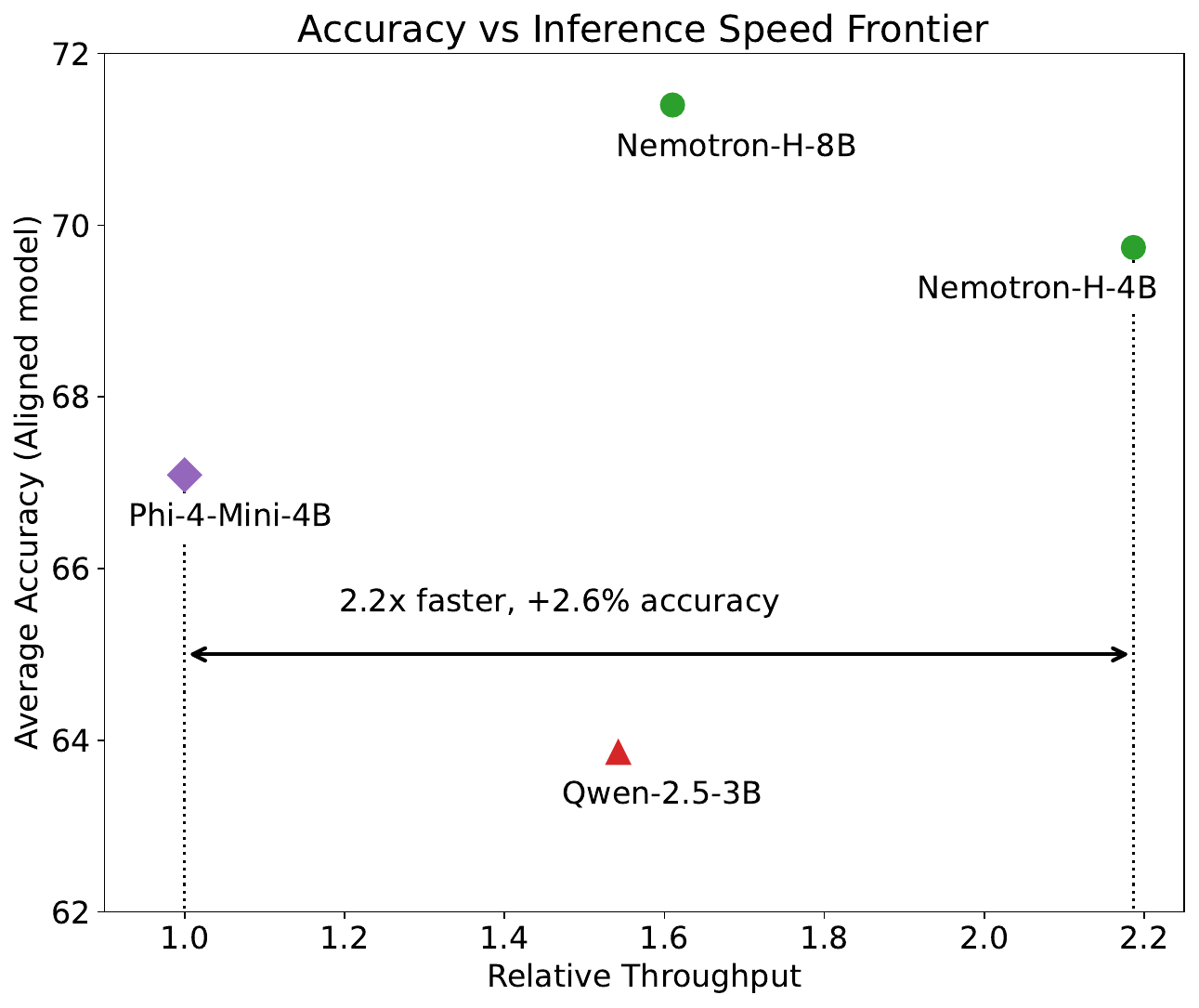}
        \label{fig:accuracy_vs_throughput}
    \end{subfigure}
    \hfill
    \begin{subfigure}[t]{0.5025\textwidth}
        \centering
        \includegraphics[width=\linewidth]{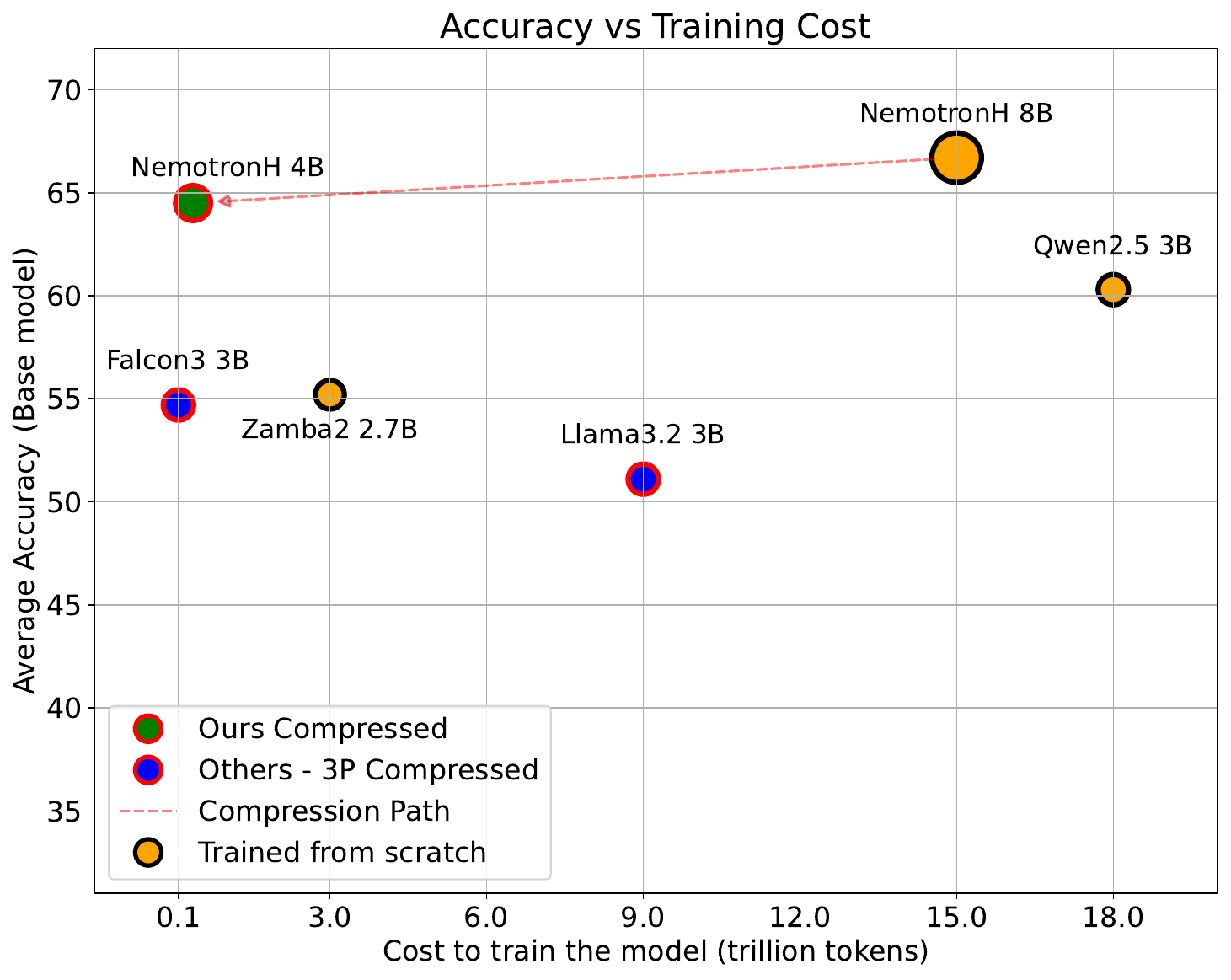}
        \label{fig:accuracy_vs_tokens}
    \end{subfigure}
    \caption{Comparison of Nemotron-H 4B model accuracy w.r.t.~inference throughput (left), and training budget for the base model (right) to similarly-sized community models. Inference throughput is measured at an input and output sequence length of 65536 and 1024, respectively.}
    \label{fig:teaser}
\end{figure*}

\section{Introduction}
\noindent Recent advances in language modeling have led to the development of hybrid architectures that combine Transformer layers~\cite{vaswani2017attention} with State Space Models (SSMs)~\cite{gu2023mamba, dao2024transformers}. These hybrid models leverage the complementary strengths of both approaches: Transformers excel at capturing global dependencies through self-attention mechanisms, while SSMs provide efficient sequence processing with $O(N)$ scaling during training and $O(1)$ cache size during inference. Mamba~\cite{gu2023mamba, dao2024transformers} in particular is a popular SSM designed for efficient sequence modeling with linear-time complexity and support for long contexts and is often the preferred choice for non-attention layers in hybrid architectures.
Despite their improved efficiency, many hybrid LLMs remain incredibly large, often spanning billions of parameters - this motivates the need for efficiently creating smaller hybrid models suitable for deployment in various resource-constrained environments.

Model pruning—the removal of redundant parameters while preserving accuracy—has recently emerged as a promising approach for compressing LLMs. In particular, methods that combine structured pruning  (i.e., pruning of entire parameter blocks such as neurons, attention heads, etc.) with knowledge distillation~\cite{hinton2015distilling} have proven effective at simultaneously reducing model memory footprint while improving runtime performance and accuracy~\cite{Minitron}.
While pruning techniques have been extensively studied for Transformer architectures~\cite{Minitron, bercovich2024puzzledistillationbasednasinferenceoptimized, tang2025darwinlm}, their application to hybrid models remains significantly underexplored. 


Some early work on Mamba and SSM pruning includes Mamba-Shredder~\cite{munoz2025mamba}, which removes the entire \textit{state space module} from the Mamba layers, leaving only linear projections and a convolution layer.
In a concurrent study, Ghattas et al.~\cite{ghattas2025pruning} proposed a method for pruning Mamba architectures by focusing on three aspects: 
state space dimension reduction, Mamba head dimension pruning, and Mamba head merging. 
To the best of our knowledge, no existing work on SSM/Mamba pruning presents a holistic compression strategy that simultaneously combines various aspects of SSM pruning with the pruning of other network components such as FFN neurons, embedding channels, and network depth; we believe such an approach is essential for obtaining the best combination of runtime performance and model accuracy.

In this paper, we introduce a novel pruning method for Mamba architectures that compresses multiple dimensions (Mamba heads, head channels). We also present a unified pruning recipe that combines Mamba pruning with FFN, embedding dimension, and layer pruning to maximize accuracy and runtime performance.
This paper makes the following key contributions:

\begin{itemize}
    \item Introduces a group-aware pruning method for Mamba layers that preserves SSM block structure and sequence modeling capabilities.
    \item Presents a novel {\em hybrid pruning recipe} that effectively combines Mamba pruning with the pruning of other network components such as FFN neurons, embedding channels and layers.
    \item Presents findings on the sensitivity of Mamba block components to pruning, along with accuracy-throughput trade-offs when combined with pruning of other network components.
    \item Utilizes the proposed hybrid pruning recipe to compress the Nemotron-H 8B model to 4B parameters through pruning and knowledge distillation. The resulting model requires up to $\sim$40x fewer training tokens compared to others in the same size range. It also achieves state-of-the-art accuracy on benchmarks, along with a $\sim$2x speedup in throughput compared to models of similar size, significantly pushing the Pareto frontier.
\end{itemize}



%% file: tex/background.tex
\section{Background}
\noindent\textbf{State Space Models (SSMs).} SSMs are a class of sequence models that process inputs through hidden states evolving over time~\cite{dao2024transformers}. The general form of an SSM is given by:

\begin{align}
    h_t &= A h_{t-1} + B x_t \label{eq:1a} \\
    y_t &= C^\top h_t + D x_t \label{eq:1b}
\end{align}

Here, \(h_t\) represents the hidden state, \(x_t\) the input, \(y_t\) the output, and \(A\), \(B\), \(C\), and \(D\) are parameter matrices. The above equations describe linear time-invariant (LTI) SSMs, where the parameters remain constant across timesteps. The Mamba architecture~\cite{dao2024transformers} introduced a selective SSM variant with time-varying parameters:

\begin{align}
    h_t &= A_t h_{t-1} + B_t x_t \label{eq:2a} \\
    y_t &= C_t^\top h_t + D_t x_t \label{eq:2b}
\end{align}

This selective mechanism allows the model to adapt dynamically to the input sequence, improving performance on complex tasks.
Mamba2~\cite{dao2024transformers} builds upon the selective SSM framework and introduces several enhancements to improve efficiency and scalability. It leverages the Structured State Space Duality (SSD), which connects SSMs and attention mechanisms through semi-separable matrix representations. This duality enables Mamba2 to combine the linear efficiency of SSMs with hardware-friendly quadratic computations typical of attention models.

\noindent\textbf{SSM-Transformer Hybrid Model}
 architectures combine State Space Models (SSMs) and Transformers to leverage complementary strengths: SSMs enable linear-scaling long-sequence processing, while transformers provide contextual reasoning. Recent implementations demonstrate this synergy—\textit{Nemotron-H}~\cite{blakeman2025nemotron} is a family of hybrid Mamba2/Transformer architectures that replaces 92\% of attention layers with constant-memory Mamba2~\cite{dao2024transformers} blocks, achieving state-of-the-art accuracy while delivering up to 3x higher inference throughput compared to pure Transformers. \textit{Jamba}~\cite{jamba2024hybrid} incorporates mixture-of-experts (MoE) modules, and cuts KV cache sizes 8×, supporting 256K-token contexts. \textit{Zamba}~\cite{glorioso2024zamba} further enhances parameter efficiency through shared global attention and low-rank projections, maintaining performance with minimal resources.
These architectures demonstrate three key advantages over pure Transformer architectures: (1) drastically reduced KV cache requirements enabling memory-efficient long-context processing, and (2) increased throughput via SSM-based sequence modeling. By balancing SSMs' computational efficiency with Transformers' expressivity, hybrid models address critical limitations in pure Transformers approaches for large-scale sequence tasks.

\noindent\textbf{Model Pruning.}
Weight pruning is a powerful and well-known technique for reducing model size~\cite{Minitron,wang2024model,hoefler:2021}. In particular, {\em structured pruning} removes blocks of nonzero elements at once from model weights, making it easier to realize actual hardware speedups; examples of structured pruning techniques include neuron, attention head, convolutional filter, and depth pruning~\cite{Minitron,luo:2017,he:2018,xia2023sheared,ashkboos2023slicegpt,men2024shortgpt,yang2024laco,kim2024shortened}. In most recent work, pruning is typically divided into three phases: (1) importance estimation, (2) model trimming, and (3) accuracy recovery. Here, importance estimation computes the importance or sensitivity of various network components (attention heads, layers, etc.). These components are then sorted in decreasing order of importance, following which the corresponding weight matrices are reshaped (trimmed). The pruned model typically loses a lot of accuracy in this process, which is then recovered using continued training. Recent work~\cite{Minitron} has demonstrated that knowledge distillation~\cite{hinton2015distilling} can be an effective alternative to traditional fine-tuning for accuracy recovery.

%% file: tex/method.tex
\section{Methodology}
\label{sec:method}

\begin{figure*}[h]
    \centering
        \includegraphics[width=\textwidth]{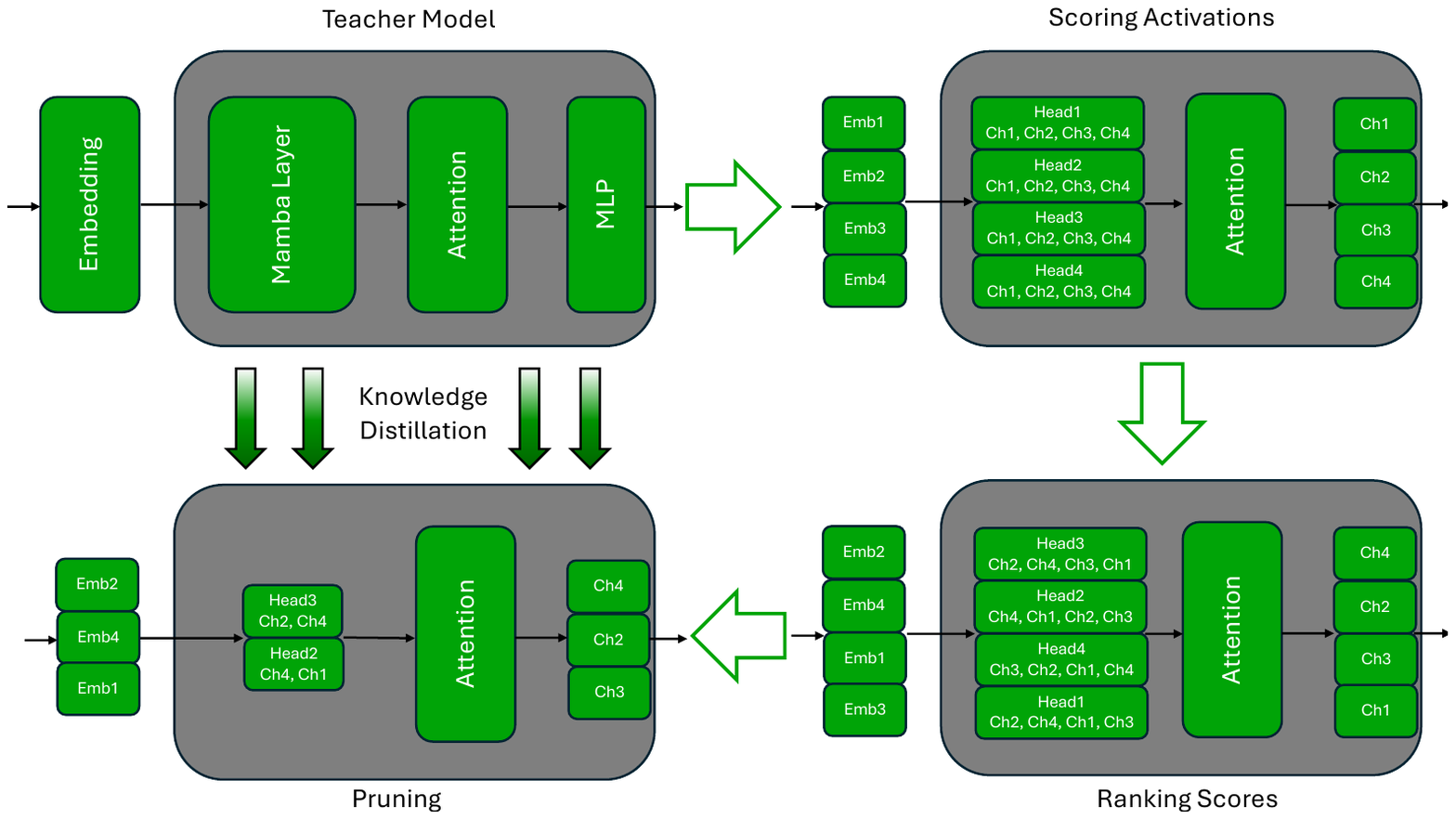}
        \caption{Overview of pruning and distillation for hybrid architectures. Starting from a pretrained LLM, we first evaluate the importance of Mamba heads and channels, FFN neurons, and embedding channels. We then rank them, trim the least important neurons, and distill the knowledge from the original LLM to the pruned model. Attention layers are not pruned since they amount to only 8\% of the total number of layers.}
        \label{fig:method}
\end{figure*}

We start the pruning procedure by computing the importance or sensitivity of each network component; namely, Mamba heads and head channels, FFN neurons, embedding channels, and layers. To keep this phase lightweight, we adopt a purely activation-based strategy (requiring only forward propagation passes) for computing importance scores, similar to Minitron~\cite{Minitron}. Once scores are computed, we sort the corresponding network components in decreasing order of importance while following any additional implementation constraints (discussed in more detail in the following subsection). We then prune away the network components with the lowest scores. Finally, the pruned model is distilled using the teacher model to obtain the final pruned model. The full procedure is illustrated in Figure~\ref{fig:method}.


\begin{figure*}[h]
        \centering
        \includegraphics[width=\textwidth]{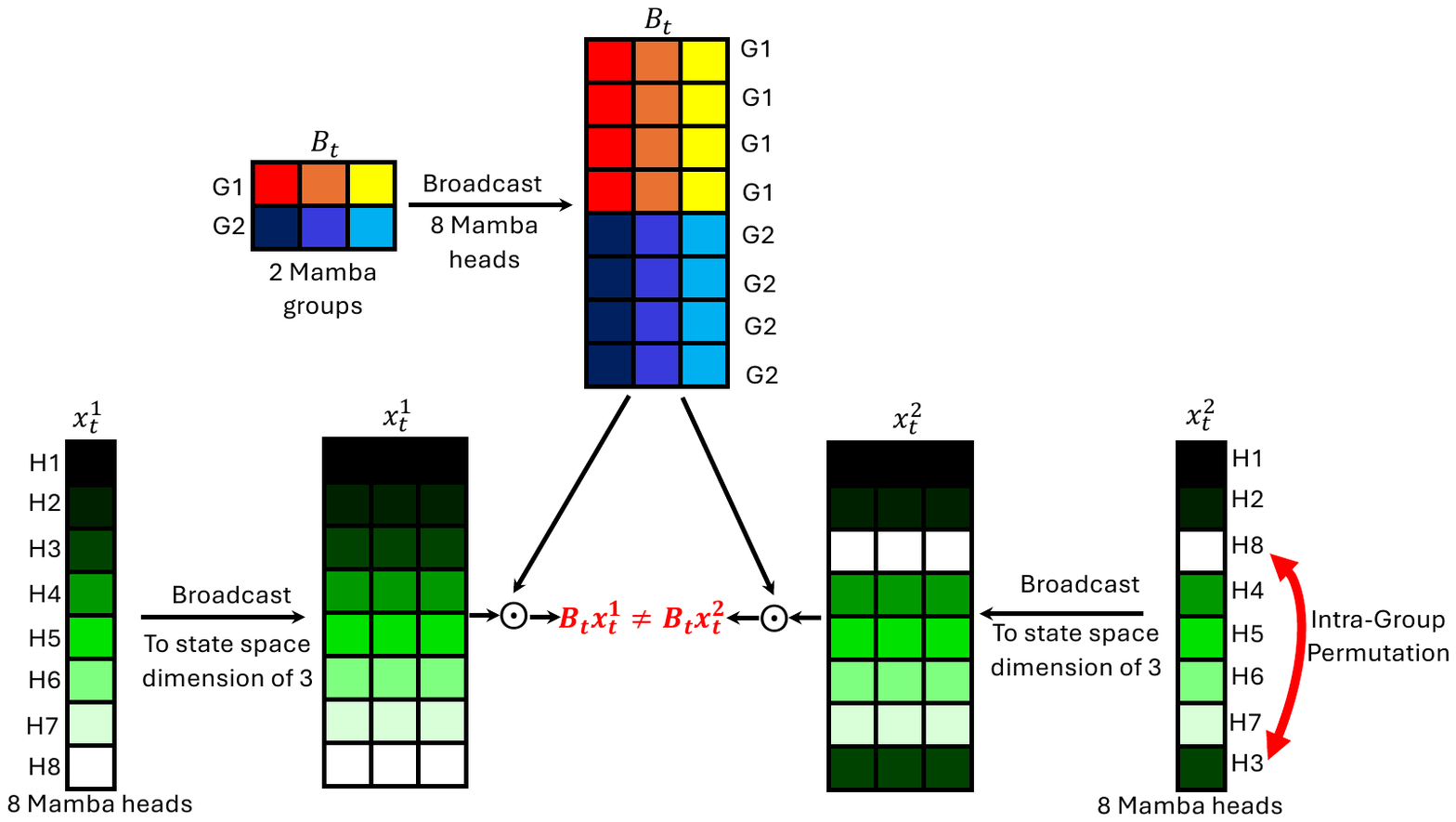}
        \caption{Mamba group structure visualization showing broadcasting and original $B_tx_t$ computation. Colors represent distinct entries. The Figure illustrates how only within-group head permutations can preserve SSM semantics. As a counter example, if H3 and H8 were to be swapped, the resulting $B_tx_t$ would NOT be any permutation of the original (no permutation) $B_tx_t$.}
        \label{fig:B_x_t_permutation}
\end{figure*}

\subsection{Mamba Pruning}
\label{subsec:mambapruning}
We now describe the importance estimation and pruning of Mamba layers in more detail. To understand the pruning procedure better, we first dive into the forward pass of a Mamba layer.

The Mamba layer processes input through five distinct projection matrices $W_z, W_x, W_B, W_C,$ and $W_{d_t}$, following layer normalization.
These projections generate intermediate matrices~\footnote{We factor out the sequence length and batch size to simplify our description; the analysis remains valid without them.}:
\begin{align}
z &= W_z(\text{LN}(X)), W_z \in \mathbb{R}^{d_e \times (m_h \times m_d)} \label{eq:wz_activation} \\
x &= W_x(\text{LN}(X)), W_x \in \mathbb{R}^{d_e \times (m_h \times m_d)} \\
B &= W_B(\text{LN}(X)), W_B \in \mathbb{R}^{d_e \times (g \times d_s)} \\
C &= W_C(\text{LN}(X)), W_C \in \mathbb{R}^{d_e \times (g \times d_s)} \\
d_t &= W_{d_t}(\text{LN}(X)), W_C \in \mathbb{R}^{d_e \times m_h}
\end{align}
Where $X$ is the layer input and LN denotes layer normalization. $d_e$ is model embedding dimension (AKA hidden dimension), $g$ is the number of Mamba groups, $d_s$ is the SSM state dimension, $m_h$ is number of Mamba heads, and $m_d$ is Mamba head channels. The matrices $x, B,$ and $C$ undergo causal convolution before participating in the selective state space model (SSM) updates:

\begin{align}
\hat{x} &= \text{conv1d}(x) \\
\hat{B} &= \text{conv1d}(B) \\
\hat{C} &= \text{conv1d}(C) \\
\tilde{y} &= \text{SSM}(\hat{x}, \hat{B}, \hat{C}, A, D, d_t)
\end{align}

Here, $A, D\in \mathbb{R}^{m_h}$ are SSM learnable parameters corresponding to state transition and direct feed through, respectively (see Equations~\ref{eq:2a} and~\ref{eq:2b}).

The SSM output is fed into a gated normalization layer, which is then followed by output projection, $W_O \in \mathbb{R}^{(m_h \times m_d) \times d_e}$: 

\begin{align}
y = W_O(\text{RMSNorm}(\tilde{y}, z)) \label{eq:wo_activation}
\end{align}

\noindent\textbf{Group-Aware Head Permutation Constraints}
Pruning requires scoring, sorting, and trimming neurons or heads of each layer, as shown in Figure~\ref{fig:method}. The FFN and embedding activations are permutation equivariant, i.e. for a permutation operator $\mathcal{P}$, FFN or embedding layer $L$, and activation  $\mathcal{A}$, and input $X$ we have:
\begin{align}
L(X) = \mathcal{A} \implies \mathcal{P}(L)(X) = \mathcal{P}(\mathcal{A}).
\end{align}

However, Mamba layers and activations are not permutation equivariant. As shown in Figure~\ref{fig:B_x_t_permutation}, the $B_tx_t$ operation from Eq.~\ref{eq:2a} involves reshaping $B$ into $B \in \mathbb{R}^{g \times d_s}$, and broadcasting it across $x \in \mathbb{R}^{(m_h \times m_d)}$. This broadcasting creates group-specific interaction patterns that constrain our pruning approach.
As a result, permuting heads across groups would alter the $B_tx_t$ broadcast pattern, violating Eq.~\ref{eq:2a}'s group-wise computation as shown by:
\begin{align}
B_tx_t &\neq (B\mathcal{P}(x_t))\mathcal{P}^T \label{eq:perm_invariant}
\end{align}

Therefore, when sorting Mamba heads using activation scores, we must preserve Mamba's group structure. Let $\mathcal{G}_g \subset \{1,...,m_h\}$ denote the set of heads belonging to group $g$. Any permutation $\mathcal{P}$ of heads must satisfy:
\begin{equation}
\mathcal{P}(h) \in \mathcal{G}_g \quad \forall h \in \mathcal{G}_g.
\label{eq:group_constraint}
\end{equation}

In other words, Mamba heads and activations are permutation equivariant only for the permutation operators defined in constraint~\ref{eq:group_constraint}.

\noindent\textbf{Head Channel Consistency.}
A similar constraint for permuting Mamba head channels applies. For head channel pruning, we maintain consistency across all heads through shared ranking. The state tensor $h \in \mathbb{R}^{m_h \times m_d \times d_s}$ requires channel-wise permutations $\mathcal{P}_d$ to satisfy:
\begin{equation}
\mathcal{P}_d(h_{i,j,k}) = \mathcal{P}_d(h_{i',j,k}) \quad \forall i,i' \in \{1,...,m_h\}
\end{equation}
meaning each channel index $k$ is either preserved or pruned uniformly across all heads.


\noindent\textbf{Scoring and Ranking Methodology.}
The Mamba head and head channel ranking follows a nested scoring procedure:

1. \textbf{Head Channel Scoring}:
For each head channel $d \in \{1,...,m_d\}$, we compute aggregate importance scores:
\begin{align}
s &= LN(X)(W_x)^{T} \label{eq:head_score} \\
s_d &= \|\sum_{B,L} s_{:,d}\|_{2}
\end{align}
where the aggregation is over $L$, the sequence length, and $B$, the batch size. Aggregation metric used along $L$ and $B$ dimensions are mean and $L_2$, respectively, following Minitron~\cite{Minitron}. $s \in \mathbb{R}^{(m_h \times m_d)}$ contains raw activation scores, and $\mathbf{s}_{:,d}$ denotes the $d$-th column across all heads. We then select the top-$k_d$ channels:
\begin{equation}
\mathcal{D}_{\text{top}} = \underset{d \in \{1,...,m_d\}}{\text{topk}}(s_d, k=k_d)
\end{equation}

2. \textbf{Head Scoring}:
Using the pruned channels $\mathcal{D}_{\text{top}}$, compute head importance scores:
\begin{equation}
f_h = \left\| \mathbf{s}_{h,\mathcal{D}_{\text{top}}} \right\|_2 \quad \forall h \in \{1,...,m_h\}
\end{equation}

3. \textbf{Group-Constrained Ranking}:
Within each Mamba group $\mathcal{G}_g$, sort heads by their scores:
\begin{equation}
\mathcal{R}_g = \underset{h \in \mathcal{G}_g}{\text{argsort}}(f_h)
\end{equation}
The final head ranking $\mathcal{R}$ is the concatenation of group-wise rankings:
\begin{equation}
\mathcal{R} = \bigoplus_{g=1}^G \mathcal{R}_g[1:k_g]
\end{equation}
where $k_g$ is the target head count per group and $\bigoplus$ denotes ordered concatenation.

The following algorithm provides a concise walkthrough on how to obtain mamba head and head channel rankings: 
\begin{algorithmic}[1]
\Require Activation scores $\mathbf{s} \in \mathbb{R}^{m_h \times m_d}$, target channels $k_d$, target heads per group $\{k_g\}_{g=1}^G$
\Ensure Head ranking $\mathcal{R}$, channel ranking $\mathcal{D}_{\text{top}}$
\State Compute channel scores: $s_d \gets \left\|\mathbf{s}_{:,d}\right\|_2\;\forall d$
\State $\mathcal{D}_{\text{top}} \gets \text{top-$k_d$ indices of } \{s_d\}$
\State Compute head scores: $f_h \gets \left\|\mathbf{s}_{h,\mathcal{D}_{\text{top}}}\right\|_2\;\forall h$
\For{$g \gets 1$ \textbf{to} $G$}
    \State $\mathcal{R}_g \gets \text{argsort-descending}(\{f_h | h \in \mathcal{G}_g\})$
    \State $\mathcal{R}_g^{\text{sel}} \gets \text{first } k_g \text{ elements of } \mathcal{R}_g$
\EndFor
\State $\mathcal{R} \gets \bigoplus_{g=1}^G \mathcal{R}_g^{\text{sel}}$ 
\end{algorithmic}

After obtaining the Mamba heads and head channel neurons to keep, we trim the corresponding matrices:

\begin{align}
W &\xleftarrow{} W[\mathcal{R}], \nonumber \\
  &\text{for } W \in \{W_x, W_z, W_O, W_A, W_D, W_{d_t}, \text{conv1d}\}
\end{align}

\subsection{FFN and Embedding Pruning}

For FFN and embedding channels, we compute importance scores using activation-based metrics. Similar to the approach in structured pruning of transformers~\cite{Minitron}, we examine the activations produced by the FFN and LayerNorm layers to determine which neurons and embedding channels contribute least to the model's performance.

For the $i$-th neuron in a feed-forward layer, we compute its importance score as:
\begin{equation}
F_{\text{neuron}}^{(i)} =  \sum_{B,L} X(W_1^i)^T \label{eq:score_ffn}
\end{equation}
where $W_1^i$ refers to the $i$-th row of the weight matrix $W_1$ in the first linear projection of the FFN, $X$ is the input to the FFN layer, and $\sum_{B,L}$ denotes aggregation along the batch and sequence dimensions. 

Similarly, for the $i$-th embedding channel, we compute:
\begin{equation}
F_{\text{emb}}^{(i)} = \sum_{B,L} \text{LN}(X)_i \label{eq:score_emb}
\end{equation}
where $\text{LN}(X)_i$ represents the $i$-th dimension of the layer-normalized input. The embedding channel scores are computed across all layers that utilize the embedding channel, including FFN, Mamba and Attention projection layers, and LayerNorm components. Aggregation metric used along $L$ and $B$ dimensions are mean and $L_2$, respectively, for both embedding Equations~\ref{eq:score_ffn}.

After computing these scores, we sort them in descending order and keep the top-k neurons and embedding channels based on the target compression ratio, pruning those with the lowest importance scores.

\subsection{FLAP Importance for Hybrid Models}

FLAP~\cite{flap} is a retraining-free structured pruning technique designed to measure the recoverability of a model's output feature map upon removing specific columns from weight matrices. FLAP quantifies the ``fluctuation'' of each input feature relative to a baseline using calibration data. Specifically, the FLAP importance score for a column is computed as the product of the squared norm of the column weights and the sample variance of the corresponding input features across calibration samples.

We extend FLAP to the SSM layers in hybrid architectures by applying the metric to the activations serving as inputs to the output projection (OutProj) matrix. Here, we compute the FLAP importance by assessing the variance in activations input to the OutProj matrix, weighted by the squared norms of the respective columns of the OutProj weights. Mathematically, the extended FLAP importance metric for a given column \( j \) of weight matrix \( W \) in SSM layers can be defined as:

\[
S_j = \|W_j\|^2 \cdot \mathrm{Var}(X_j)
\]

where \( \|W_j\|^2 \) denotes the squared norm of the column weights and \( \mathrm{Var}(X_j) \) represents the variance of the activations input to the output projection matrix of SSM layer across calibration samples.

We use the above-computed metric to rank different heads within each group and remove the corresponding rows in the input projection matrix, the corresponding channels in the SSM convolution kernel, corresponding rows in the \( A \) and \( D \) matrices of SSM, as well as trimming the corresponding columns in the output projection matrix.

\subsection{Depth Pruning}
We explored depth pruning by analyzing layer importance using Kullback-Leibler divergence (KLD) between logits from a model with a specific layer removed and the full model. This importance estimation was averaged over a small random subset of 256 samples to account for sample variability.

\begin{figure*}[h!]
    \centering
    \begin{minipage}{0.48\textwidth}
        \centering
        \includegraphics[width=\textwidth]{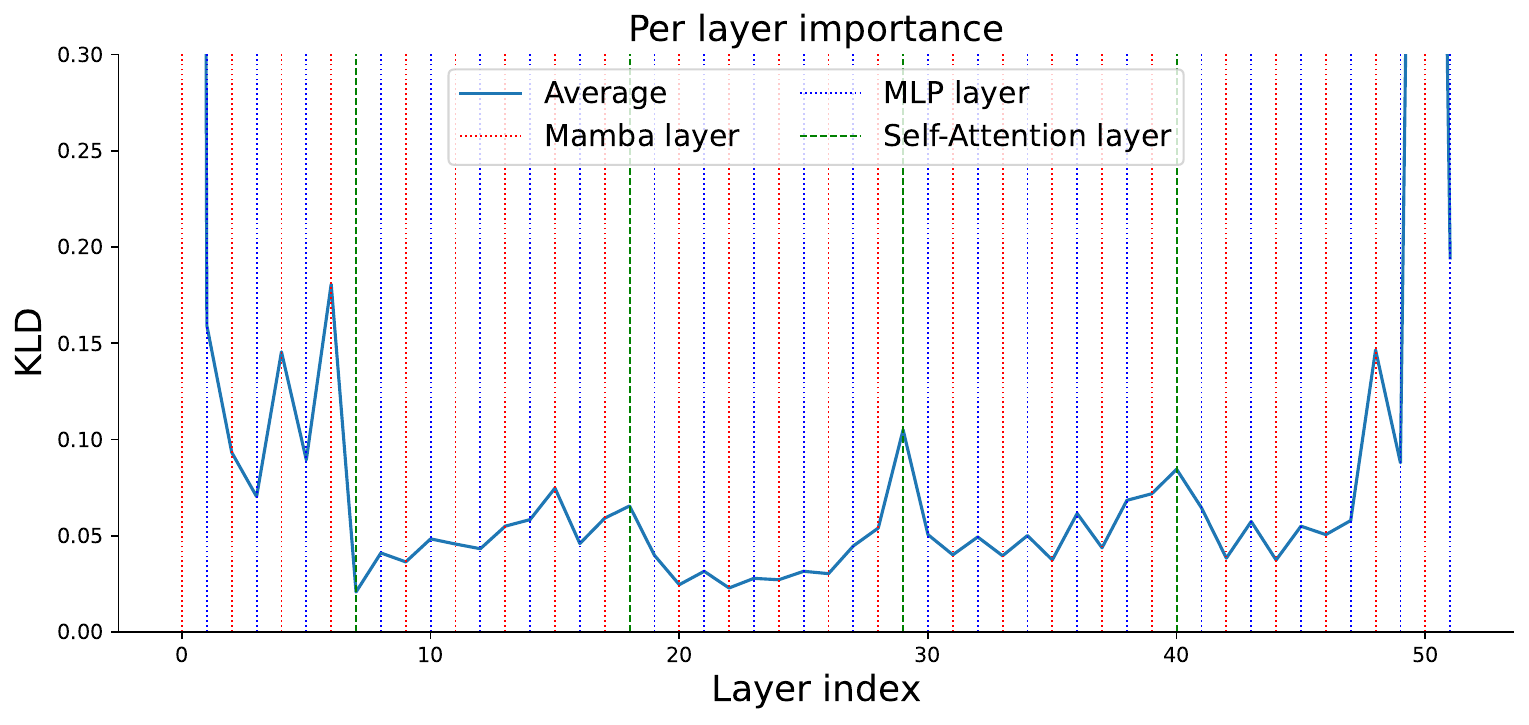}
        \caption{Layer importance measured as the KLD between logits of the full model and a model with that layer removed, averaged over a small training subset. Vertical dotted lines indicate layer types: self-attention (green), FFN (blue), and Mamba2 (red).}
        \label{fig:per-layer-importance}
    \end{minipage}%
    \hfill
    \begin{minipage}{0.48\textwidth}
        \centering
        \includegraphics[width=\textwidth]{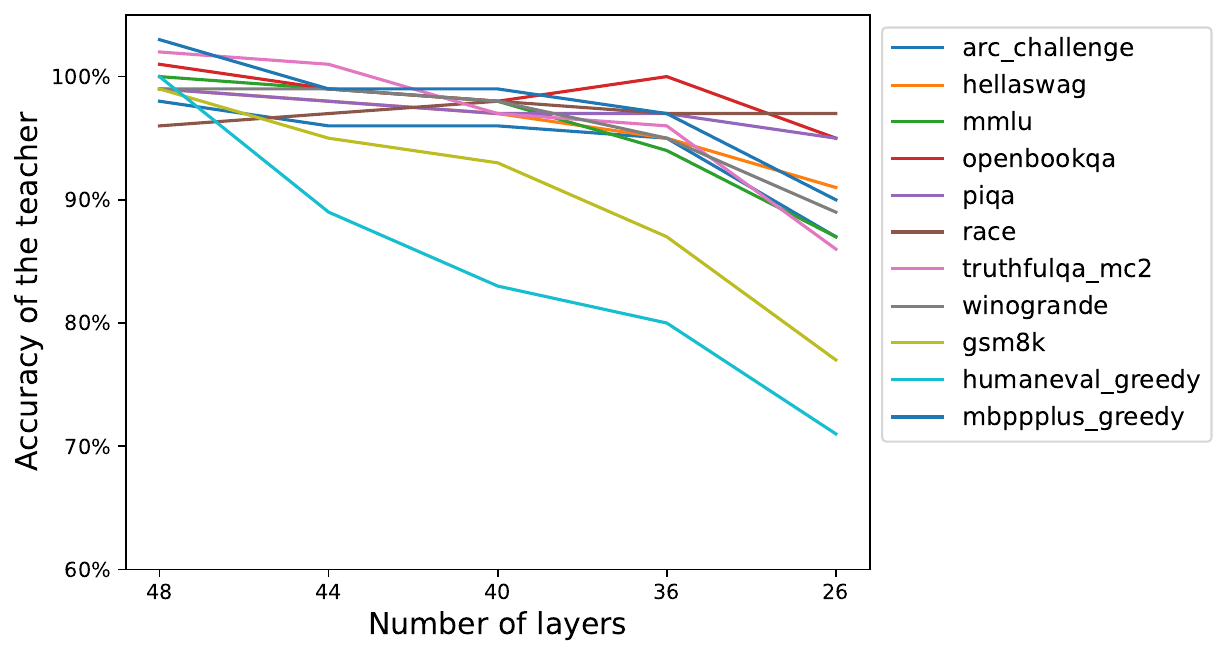}
        \caption{Accuracy drop relative to the 8B model across progressively depth-only pruned variants (48, 44, 40, 36, and 26 layers). Each model is directly pruned from the 8B and distilled using 126B tokens.}
        \label{fig:iterative-layer-removal}
    \end{minipage}
\end{figure*}

Figure \ref{fig:per-layer-importance} shows the average importance scores for each layer in the Nemotron-H 8B Base model, with green, blue, and red dotted lines representing self-attention, FFN, and Mamba layers. As seen in previous work~\citep{blakeman2025nemotron}, the most important layers are concentrated at the model's start and end. Interestingly, the first attention layer is among the least important, while other attention layers are more critical than neighboring layers. A ``saw-like'' pattern emerges where MLP layers are more important than adjacent Mamba layers in the middle of the network, though this reverses in the model's critical regions.

We experimented by pruning the least important layers (4, 8, 12, 16, and 26 layers), followed by distillation with 126B tokens. While core-knowledge benchmarks remained largely unaffected, tasks like math and coding showed significant performance degradation (Figure \ref{fig:iterative-layer-removal}).


\subsection{Architecture Search}
\label{subsec:search}

Our compression strategy explores multiple axes within the 4B parameter budget through combinatorial pruning. Our search space includes depth reduction (removing 4-26 layers from the original 52-layer architecture) combined with width pruning of embedding channels (3072-4096), FFN dimension (9984-21504), Mamba heads (64-128), and Mamba head channels (32-64). This multi-axis search space generated over a hundred candidate architectures meeting the parameter constraints.

Our search procedure follows these steps: (1) compute the zero-shot validation loss for all candidates on 1024 calibration samples, (2) select the top K architectures (22 in this study) with the best loss values and perform lightweight knowledge distillation (KD) on them with 3.8B tokens, using the original 8B model as the teacher, and (3) select the top architecture candidate from step (2), using throughput and latency measurements for breaking ties, and perform extended knowledge distillation with $\sim380B$ tokens to obtain the final model (see Table~\ref{tab:base_model_results}). We note that step (2) is critical for getting a reliable ranking of architectural candidates, as also noted in prior work~\cite{Minitron}.

\subsection{Accuracy Recovery with Knowledge Distillation (KD)}
To recover the accuracy lost due to pruning, the model undergoes continued training. Recent work has demonstrated that distilling knowledge~\cite{hinton2015distilling} from the original model to the pruned model outperforms conventional fine-tuning~\cite{sreenivas2024llmpruningdistillationpractice,bercovich2024puzzledistillationbasednasinferenceoptimized};
we thus adopt logit-based distillation for continued training, employing forward KL divergence (FKLD) loss exclusively during the accuracy recovery phase.

The output probability distribution of an LLM for a given token $x_i$ is computed as:
$
p(x_i, \tau) = \frac{\exp\left(\frac{x_i}{\tau}\right)}{\sum_{j=1}^{|V|} \exp\left(\frac{x_j}{\tau}\right)}$,
where $\tau$ is the softmax temperature and ${|V|}$ is the vocabulary size. Logit-based KD loss across the sequence of all output tokens is represented as:
$
L_{\text{logits}} = \frac{1}{L}\sum_{k=1}^L \text{FKLD}(p_t^k(x, \tau), p_s^k(x, \tau))
$; here, $p_t^k(x, \tau)$ and $p_s^k(x, \tau)$ represent the teacher and student probability distributions on the $k^{th}$ token, respectively, and $L$ represents the sequence length.


%% file: tex/results.tex
\section{Experiments and Results}
\label{sec:experiments}

\begin{table}[h!]
\centering
\small
\resizebox{1.0\linewidth}{!}{
\rowcolors{2}{white}{gray!10}
\begin{tabular}{c ccccccc}
\toprule
\textbf{\#} & \textbf{Layers} & \textbf{Emb} & \textbf{FFN} & \textbf{Heads} & \textbf{Head Channel} & \textbf{LM Val Loss} & \textbf{Relative Throughput} \\
\midrule
\rowcolor{yellow!30}
\textbf{1} & \textbf{52} & \textbf{3072} & \textbf{12288} & \textbf{112} & \textbf{64} & \textbf{1.380} & \textbf{1} \\
2 & 52 & 3072 & 10752 & 128 & 64 & 1.380 & 0.98 \\
3 & 52 & 3328 & 9984  & 112 & 64 & 1.384 & 1 \\
4 & 52 & 3072 & 12288 & 112 & 60 & 1.388 & 1.02 \\
5 & 52 & 3072 & 12288 & 120 & 56 & 1.388 & 1.01 \\
6 & 52 & 3072 & 13056 & 112 & 56 & 1.389 & 1.04 \\
7 & 44 & 3072 & 14592 & 128 & 64 & 1.393 & 1.11 \\
8 & 44 & 3584 & 10752 & 120 & 64 & 1.394 & 1.12 \\
9 & 52 & 3072 & 11520 & 112 & 64 & 1.396 & 1.02 \\
10 & 52 & 3072 & 13056 & 96  & 64 & 1.396 & 1.04 \\
11 & 52 & 3072 & 13824 & 128 & 48 & 1.396 & 1.03 \\
12 & 52 & 3072 & 12288 & 104 & 62 & 1.397 & 1.03 \\
13 & 52 & 3072 & 13056 & 104 & 60 & 1.397 & 1.03 \\
14 & 52 & 3072 & 13056 & 96  & 62 & 1.397 & 1.04 \\
15 & 52 & 3072 & 14592 & 96  & 56 & 1.398 & 1.05 \\
16 & 48 & 3072 & 12288 & 128 & 64 & 1.398 & 1.08 \\
17 & 48 & 3328 & 9984  & 128 & 64 & 1.399 & 1.07 \\
18 & 52 & 3072 & 13824 & 96  & 58 & 1.401 & 1.05 \\
19 & 52 & 3072 & 11520 & 128 & 56 & 1.402 & 1.01 \\
20 & 44 & 3328 & 11648 & 128 & 64 & 1.402 & 1.12 \\
21 & 48 & 3072 & 13824 & 112 & 64 & 1.403 & 1.09 \\
22 & 48 & 3328 & 11648 & 112 & 64 & 1.403 & 1.08 \\
23 & 52 & 3072 & 16128 & \textcolor{red}{64}  & 64 & \textcolor{red}{1.411} & 1.07 \\
24 & \textcolor{red}{26} & 4096 & 21504 & 128 & 64 & \textcolor{red}{1.533} & 1.31  \\
25* & \textcolor{red}{36} & 4096 & 21504 & 128 & 64 & \textcolor{red}{1.430} & 1.12 \\
\midrule
8B parent & 52 & 4096 & 21504 & 128 & 64 & - & 0.74 \\
\bottomrule
\end{tabular}}
\caption{Model configurations with their corresponding LM validation loss after lightweight KD (sorted in increasing order), and relative inference throughput. Highlighted row shows the best (lowest) loss. All models have $\sim4B$ parameters, except entries marked with *, which have more.}
\label{tab:short_kd_ablations}
\end{table}

To identify the optimal compression strategy for hybrid models, we conduct several ablation studies evaluating the impact of pruning different components on accuracy and inference speed. Our experiments reveal key insights and highlight differences from Transformer-only compression~\cite{Minitron}, as detailed in the following paragraphs.

\paragraph{Depth-only vs Width-only Pruning.} As shown in Table~\ref{tab:short_kd_ablations}, width-only pruning (\#1) significantly outperforms depth-only pruning (\#24) at a 50\% compression ratio (8B to 4B). Notably, a depth-pruned model with 36 layers (\#25), despite having $\sim$1.4× more parameters performs worse than the least accurate width-only pruned 4B candidate (\#23, with 64 Mamba heads), demonstrating the critical role of depth in maintaining accuracy as also observed with Transformer-only models.

\begin{figure*}[h!]
    \centering
    \begin{minipage}[t]{0.49\textwidth}
        \centering
        \includegraphics[width=\textwidth]{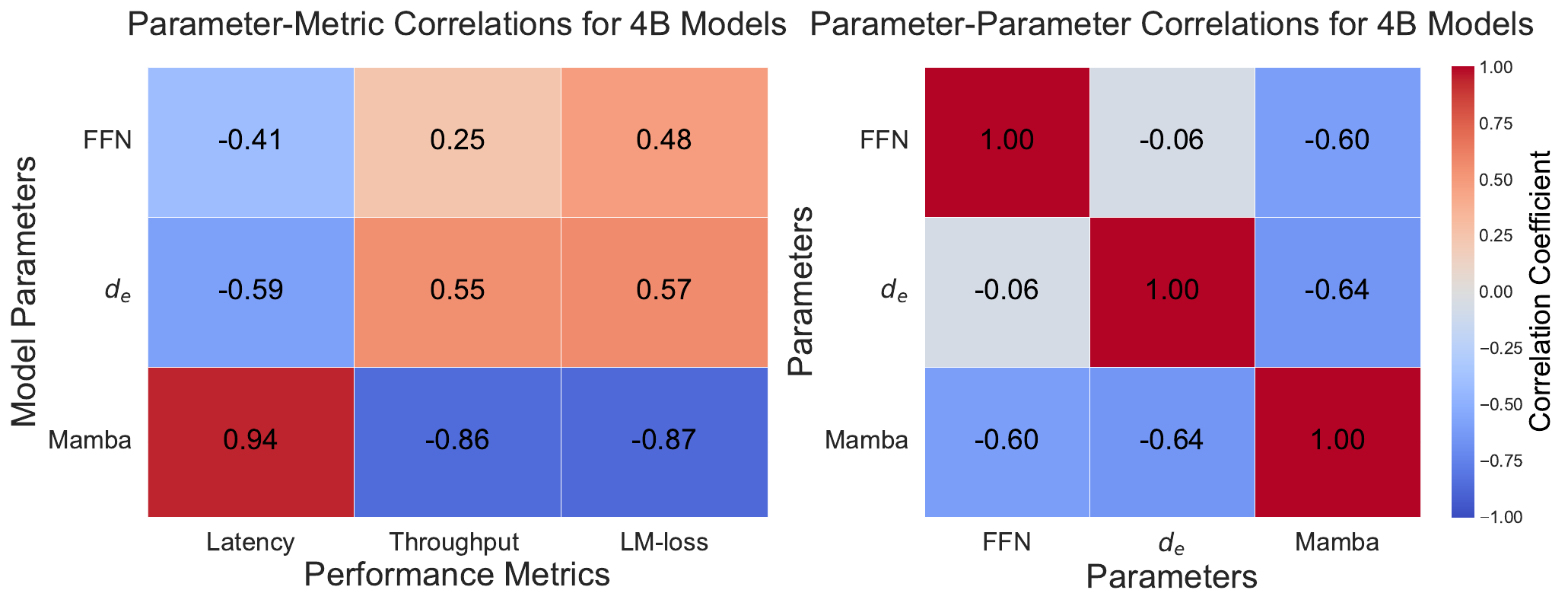}
        \caption{\textbf{Left:} Correlation matrix showing relationships between performance metrics and model components—FFN, embedding dimension ($d_e$), and Mamba parameters (varying both heads $m_h$ and head dimension $m_d$)—across 125 4B variants with fixed depth (52 layers). \textbf{Right:} Model parameter correlations for a fixed 4B parameter budget—highlighting trade-offs where increasing one component reduces others.}
        \label{fig:correlation_matrix}
    \end{minipage}
    \hfill
    \begin{minipage}[t]{0.48\textwidth}
        \centering
        \includegraphics[width=\textwidth]{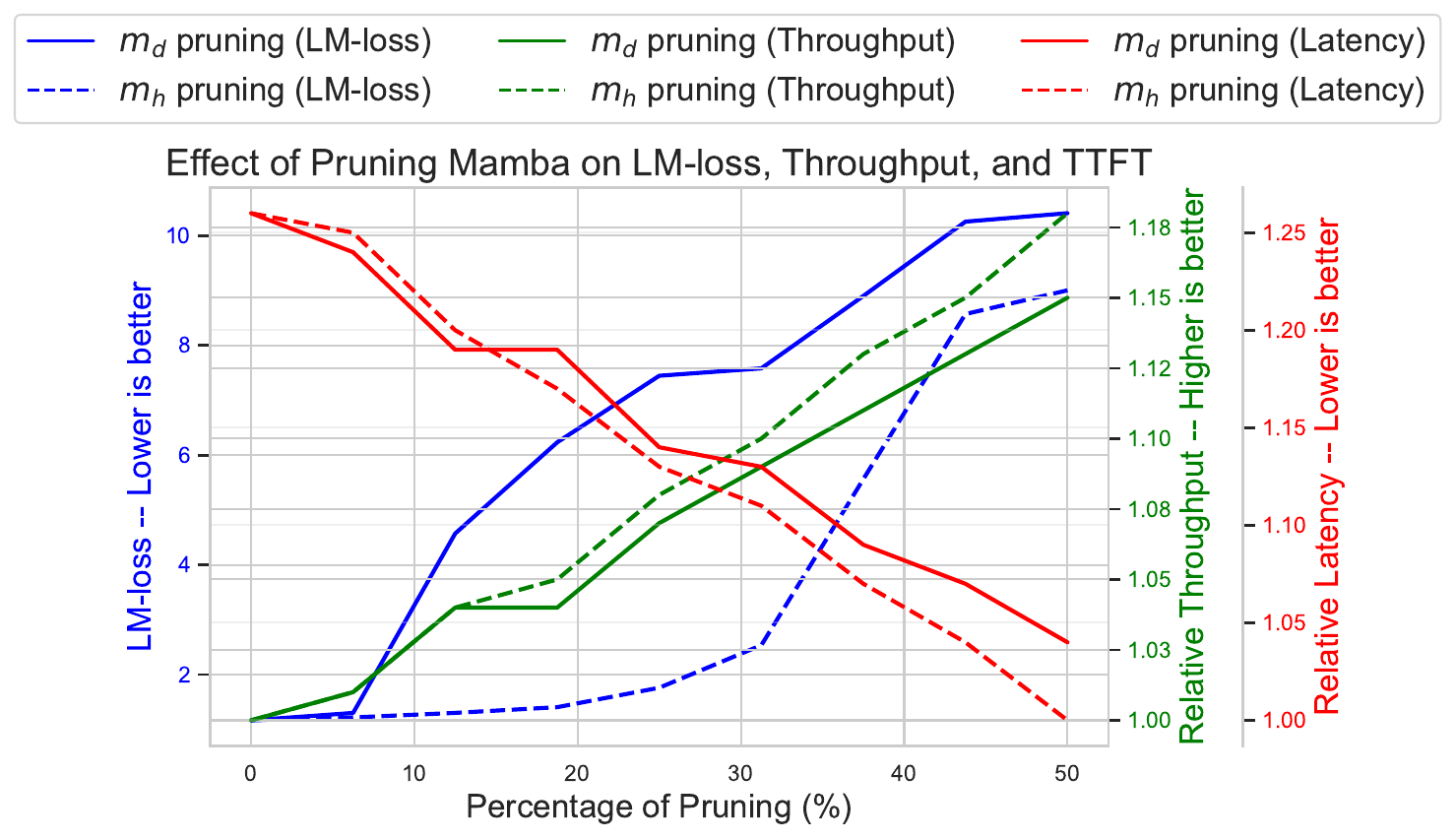}
        \caption{Impact of pruning Mamba heads ($m_h$) versus Mamba head channels ($m_d$) in isolation, with the rest of the network unchanged. Pruning $m_h$ consistently outperforms $m_d$ pruning across LM loss, latency, and throughput—establishing it as the preferred target for optimization.}
        \label{fig:mamba_heads_vs_channels}
    \end{minipage}
\end{figure*}

\paragraph{Impact on Inference Speed.} Table~\ref{tab:short_kd_ablations} shows that depth-only pruning (\#24) provides the highest speedups. Figure~\ref{fig:correlation_matrix} presents the correlation between pruning various network components and performance metrics such as throughput, latency, and LM-loss for a fixed 4B parameter count. We notice from the Figure that pruning Mamba components results in faster models compared to pruning FFN and embedding dimensions. Furthermore, we also compare the effects of pruning Mamba heads to pruning head channels in Figure~\ref{fig:mamba_heads_vs_channels}; we observe that the former yields better speed improvements than the latter within a given Mamba layer.

\paragraph{Impact on Accuracy.} Table~\ref{tab:short_kd_ablations} shows that model depth (\#24) is most sensitive to accuracy, followed by Mamba heads (\#23), while FFN and embedding dimensions have less impact. Further ablations isolating the pruning of Mamba heads and head channels show that pruning head channels leads to a greater accuracy loss (Figure~\ref{fig:mamba_heads_vs_channels}). Given depth pruning’s effect on inference speed, we explore a combined pruning strategy, starting with depth-only pruning followed by distillation to assess its limits. As shown in Figure~\ref{fig:iterative-layer-removal}, we observe significant accuracy drops on math and coding benchmarks below 44 layers. We then apply width pruning to both the 44- and 48-layer variants to produce corresponding $\sim$4B-sized models. However, we notice that the best depth-width pruned candidate (\#7, 44 layers) still under-performs the width-only model (\#1).


\paragraph{Mamba Scoring Ablations.}
In Equation~\ref{eq:head_score}, we chose the activations obtained from $W_x$ matrix for scoring the Mamba heads and head channels. We can alternatively get the Mamba scores by considering the activations obtained from $W_z$ and $W_O$ matrices, from Equations~\ref{eq:wz_activation} and~\ref{eq:wo_activation}. Table~\ref{tab:hook_ablation} shows the effect of selecting the Mamba activations from different parts of the Mamba layer. For different configurations, we notice that scoring the activations from $W_x$ output often results in the best LM loss.

\begin{table}[h!]
\centering
\small
\resizebox{1.0\linewidth}{!}{
\begin{tabular}{cccccccc}
\toprule
\textbf{FFN} & \textbf{Embedding Dim} & \multicolumn{2}{c}{\textbf{Mamba}} & \multicolumn{3}{c}{\textbf{LM-Loss}} \\
\cmidrule(lr){3-4} \cmidrule(lr){5-7}
& & \textbf{Heads} & \textbf{Head Channels} & \textbf{$W_x$} & \textbf{$W_z$} & \textbf{$W_O$} \\
\midrule
12,288 & 3,072 & 112 & 64 & \textbf{3.56} & 4.11 & 3.79 \\
13,056 & 3,072 & 112 & 56 & \textbf{3.59} & 6.61 & 5.30 \\
13,056 & 3,072 & 96 & 64 & \textbf{4.49} & 5.39 & 4.49 \\
14,592 & 3,072 & 96 & 56 & \textbf{4.68} & 7.09 & 10.01 \\
12,288 & 3,072 & 128 & 56 & 5.98 & 5.43 & \textbf{4.99} \\
13,824 & 3,072 & 128 & 48 & \textbf{5.99} & 6.01 & 9.47 \\
\bottomrule
\end{tabular}}
\caption{Mamba scoring ablation. The zero-shot LM-loss for top 6 pruned models based on Mamba scores calculated from activations of $W_x$, $W_z$, and $W_O$. The $W_x$ activations result in the best zero-shot LM-loss in most of the cases.}
\label{tab:hook_ablation}
\end{table}

\paragraph{Effect of Parameter Choice on Performance Metrics.}
In our neural architecture search, we imposed a constraint to generate valid checkpoints with a fixed size of 4 billion (4B) parameters. Within this constraint, we varied the sizes of the feed-forward network (FFN), embedding dimensions, $m_h$ (Mamba heads), and $m_d$ (Mamba head channels). As a result, we obtained 125 checkpoints, all with 4B parameters. For each checkpoint, we evaluated the lm-loss, time to first token, and throughput. To analyze the relationships between model parameters and performance metrics, we computed correlations and visualized them in Figure~\ref{fig:correlation_matrix}. Additionally, since all 125 models have the same total parameter count (4B), the model parameters exhibit negative correlations with one another.

Figure~\ref{fig:correlation_matrix} shows that in 4B models derived from Nemotron-H 8B, Mamba components positively correlate with latency and negatively with throughput and LM loss—indicating that pruning them improves inference speed and slightly degrade accuracy. In contrast, pruning embedding and FFN dimensions improves accuracy (lower LM loss) but leads to slower models with increased latency and reduced throughput.

\paragraph{Closer Look at Mamba Pruning.}
We analyze the sensitivity of two axes in the Mamba layer—Mamba heads ($m_h$) and Mamba head channels ($m_d$)—to various metrics, including accuracy, latency, and throughput. In this study, each axis was pruned in isolation while keeping the rest of the network unchanged, preserving the architecture of the Nemotron-H 8B model. The objective was to determine which axis is more favorable for optimization.
As shown in Figure~\ref{fig:mamba_heads_vs_channels}, pruning Mamba heads ($m_h$) consistently outperforms pruning Mamba head channels ($m_d$) across all metrics. Specifically, reducing $m_h$ consistently yields lower LM loss, reduced latency, and higher throughput, making Mamba heads a particularly impactful and practical target for pruning. These findings emphasize the importance of selecting the appropriate axis for pruning when optimizing Mamba layers to balance computational efficiency and model performance.

\paragraph{FLAP.} Table~\ref{tab:flap-comparison} shows that FLAP-based importance estimation yields mixed results before lightweight KD across pruning strategies. After KD, it performs on par with the L2-based approach when applied to candidate \#1; it doesn't seem to offer any clear advantage, however.

\begin{table}[h!]
\centering
\resizebox{\columnwidth}{!}{
\begin{tabular}{@{}llcc@{}}
\toprule
\textbf{Pruning Type} & \textbf{Configuration} & \textbf{L2 LM Loss} & \textbf{FLAP LM Loss} \\ \midrule
Baseline & No pruning & 1.168 & 1.168 \\[2pt]
\midrule
\multirow{3}{*}{FFN} 
 & FFN = 16384 & 1.364 & \textbf{1.32} \\
 & FFN = 11568 & 1.803 & \textbf{1.64} \\
 & FFN = 8192 & 2.281 & \textbf{1.95} \\[2pt]
 \midrule
Attention & ATT Heads = 16 & \textbf{1.282} & 1.40 \\[2pt]
\midrule
\multirow{3}{*}{Mamba (SSM)} 
 & Mamba Heads = 112 & \textbf{1.305} & 1.73 \\
 & Mamba Heads = 96 & \textbf{2.150} & 4.59 \\
 & Mamba Heads = 64 & \textbf{9.040} & 11.21 \\ 
\midrule
\midrule
\multirow{2}{*}{Mixed}
 & \#1 & \textbf{3.690} & 5.854 \\
 & \#1 + Lightweight KD & \textbf{1.380} & \textbf{1.380} \\
\bottomrule
\end{tabular}}
\caption{LM loss comparison when pruning different model components using L2 and FLAP metrics. Baseline: 128 Mamba heads, 21,504 FFN size, 32 attention heads.}
\label{tab:flap-comparison}
\end{table}

\paragraph{Summary of Ablations.}
These findings highlight the importance of choosing the right pruning axes in hybrid models to balance accuracy and efficiency. Unlike Transformer-only models—where pruning attention heads is less common~\cite{Minitron}—hybrid architectures like those with Mamba layers can tolerate some head pruning, as seen with candidates \#1 and \#2 in Table~\ref{tab:short_kd_ablations}. This tolerance may stem from Mamba layers having significantly more heads (128) than self-attention layers (32).




\subsection{Obtaining the Best Compressed Hybrid Model}

For our final model, we focus on width-only pruning to prioritize accuracy, avoiding depth reduction. This choice is motivated by Nemotron-H 8B’s already compact architecture, consisting of 52 layers that include Mamba, FFN, and Attention blocks—fewer than the 64 alternating Attention and FFN layers found in comparable models like Phi-4-4B.

Based on the lightweight KD results in Table~\ref{tab:short_kd_ablations}, we select the candidate with the lowest LM validation loss. Although both candidates \#1 and \#2 have identical losses, candidate \#1 is chosen for extended KD with 380B tokens due to its higher inference throughput, enabled by the reduction in Mamba heads.

\subsection{Data and Training Hyperparameters}
We use a random sample from the Phase 3 data mixture employed for training Nemotron-H models~\cite{blakeman2025nemotron} for both importance estimation and KD.
For importance estimation, we use 1024 samples with a sequence length of 8192. For KD, the batch size is 768, with a sequence length of 8192, a cosine decay learning rate schedule (starting at 1.6e-4 and decaying to 8e-4), with a 60-step linear warmup.

\subsection{Alignment and Long Context Extension}
We perform Supervised Fine-tuning with Knowledge Distillation (SFT-KD)\footnote{https://developer.nvidia.com/blog/data-efficient-knowledge-distillation-for-supervised-fine-tuning-with-nvidia-nemo-aligner} using the Nemotron-H 8B aligned model as the teacher, along with Reward-aware Preference Optimization (RPO) \citep{nvidia2024nemotron4340btechnicalreport} and NeMo-Aligner \citep{shen2024nemoalignerscalabletoolkitefficient}. The Nemotron-H 4B base model is fine-tuned using supervision from the top-k (100) logits of the teacher over two rounds of SFT-KD: the first round uses math and coding data, while the second round focuses on instruction-following and general chat data. The instruction-tuned model is then further aligned with two rounds of RPO.

To extend the context length of the aligned Nemotron-H 4B model, we perform SFT using data designed for long-context understanding. The training data is derived by manipulating the general domain chat dataset from the second SFT-KD round during alignment. We concatenate conversation turns and introduce long-range dependencies by placing related turns far apart within the extended context. The context length is varied randomly between 128k and 512k tokens, ensuring the model learns to maintain coherence and understanding across longer sequences, enhancing its ability to process information beyond shorter context windows. We plan to explore KD for context extension as future work.

\begin{table*}[h!]
    \centering
    \resizebox{0.9\textwidth}{!}{%
    \begin{tabular}{l|cccc|c||g}
        \textbf{Benchmarks (shots)} & 
        \textbf{Llama-3.2} & 
        \textbf{Falcon-3} & 
        \textbf{Zamba-2} & 
        \textbf{Qwen-2.5} & 
        {\textbf{Nemotron-H}} & 
        {\textbf{Nemotron-H}} \\
    
        & 
        \textbf{3B-Base} & 
        \textbf{3B-Base} & 
        \textbf{2.7B-Base} & 
        \textbf{3B-Base} & 
        \textbf{4B-Base}  & 
        \textbf{8B-Base}  \\
        \midrule
ARC Challenge (0)      & 46.5 & 47.4 & 51.5 & 47.3 & \textbf{54.4} & 60.1\\ 
ARC Easy (0)           & 72.0 & 72.4 & 79.5 & 72.7 & \textbf{81.6} & 83.6\\ 
CommonsenseQA (0)      & 66.5 & 64.4 & 76.2 & \textbf{77.1} & 70.2 & 72.7 \\ 
GSM8K (8)              & 27.1 & 66.5 & 55.0 & \textbf{75.2} & 69.6 & 77.9 \\ 
HellaSwag (0)          & 74.1 & 65.3 & 76.6 & 73.6 & \textbf{77.0} & 81.2\\ 
HumanEval (0, pass@1)  & 26.8 & 39.6 & 25.0 & 37.8 & \textbf{59.8} & 57.3 \\ 
HumanEval+ (0, pass@1) & 24.4 & 32.3 & 21.3 & 33.5 & \textbf{55.5} & 53.7\\ 
MBPP (3, pass@1)       & 42.0 & 52.1 & 36.2 & 59.9 & \textbf{65.0} & 66.9\\ 
MBPP+ (0, pass@1)      & 40.7 & 40.7 & 32.8 & 50.0 & \textbf{61.1} & 58.7 \\ 
MMLU (5)               & 56.3 & 56.7 & 56.8 & 65.6 & \textbf{68.1} & 72.7 \\ 
OpenbookQA (0)         & 41.4 & 39.4 & \textbf{46.4} & 42.2 & 44.2 & 47.2 \\ 
PIQA (0)               & 78.0 & 75.5 & \textbf{80.4} & 78.8 & 79.4 & 82.2 \\ 
RACE v.3 (0)           & 66.7 & 69.7 & 73.7 & \textbf{84.5} & 80.9 & 84.0 \\ 
Social IQA (0)         & 46.8 & 45.1 & \textbf{51.8} & 49.8 & 45.1 & 45.8\\ 
TruthfulQA MC2 (0)     & 39.3 & 45.6 & 45.8 & 49.0 & \textbf{49.4} & 49.8\\ 
Winogrande (0)         & 69.5 & 65.0 & 74.3 & 68.4 & \textbf{71.3} & 76.3 \\ 
\midrule
Average                & 51.1 & 54.7 & 55.2 & 60.3 & \textbf{64.5} & 66.7 \\ 
Tokens                 & 9T & 0.1T & 3T & 18T & 0.38T & 15T \\
\bottomrule
    \end{tabular}%
    }
    
    \caption{
        Accuracy comparison of our compressed Nemotron-H 4B with other similarly sized base community models.}
    \label{tab:base_model_results}
\end{table*}

\begin{table*}[h!]
    \centering
    \resizebox{0.9\textwidth}{!}{
    \begin{tabular}{l|ccccc|c||g}
        \textbf{Benchmarks (shots)} &  
        \textbf{Phi-4-Mini} & 
        \textbf{Qwen-2.5} & 
        \textbf{Llama-3.2} &
        \textbf{Falcon-3} & 
        \textbf{Zamba-2} &
        {\textbf{Nemotron-H}} &
        {\textbf{Nemotron-H}} \\
    
        & \textbf{4B-Instruct-128k} & \textbf{3B-Instruct-32k} & \textbf{3B-Instruct-128k} & \textbf{3B-Instruct-32k} & \textbf{2.7B-Instruct-4k}  & \textbf{4B-Instruct-128k} & \textbf{8B-Instruct-128k}  \\
        \midrule
        
        MMLU (0, generative)       & 61.88           & 63.25 & 57.36 & 54.27 & 55.32 & \textbf{66.96} & 68.7 \\
        GSM8K (0)                  & 87.71           & 83.32 & 78.47 & 77.86 & 66.26 & \textbf{88.93} & 90.4 \\
        MATH-500 (0)               & 70.8            & 65.6  & 48.2  & 48.80 & 29.40 & \textbf{76.4}  & 77.6 \\
        HumanEval (0, pass@1)      & 73.17           & 75.0  & 55.49 & 46.34 & 37.20 & \textbf{76.2}  & 79.3 \\
        HumanEval+ (0, pass@1)     & 64.63           & 70.12 & 51.83 & 43.29 & 32.93 & \textbf{70.85} & 74.4 \\
        MBPP (0, pass@1)           & 67.46           & 67.72 & 65.61 & 61.37 & 46.30 & \textbf{78.6}  & 81   \\
        MBPP+ (0, pass@1)          & 60.31           & 58.47 & 55.29 & 55.03 & 38.62 & \textbf{68.25} & 67.7 \\
        IFEval Strict (0)          & 74.78           & 64.06 & 74.51 & 68.49 & 46.99 & \textbf{76.24} & 78.6 \\
        MT-Bench (0)               & 7.86            & 7.68  & 7.09  & 7.10  & 7.02  & \textbf{7.90}  & 7.90 \\
        BFCL v2 Live (0)           & 61.64           & 59.08 & 49.58 & 52.80 & 39.70 & \textbf{65.88} & 62.6 \\
        \bottomrule
    \end{tabular}%
    }   
    \caption{
        Accuracy comparison for instruction-tuned models.
        For IFEval, we report the average of prompt strict and instruction strict categories. 
        For BFCL v2, we report live overall accuracy.
        For MT-Bench, we use GPT-4-Turbo as the judge.
    } 
    \label{tab:alignment_results}
\end{table*}

\begin{table*}[h!]
    \centering
    \resizebox{0.9\textwidth}{!}{%
    \begin{tabular}{l|ccc|c||g}
        \textbf{Context Length} & 
        \textbf{Phi-4-Mini} & 
        \textbf{Qwen-2.5} & 
        \textbf{Llama-3.2} &
        {\textbf{Nemotron-H}} &
        {\textbf{Nemotron-H}} \\
    
        & \textbf{4B-Instruct-128k} & \textbf{3B-Instruct-32k} & \textbf{3B-Instruct-128k} & \textbf{4B-Instruct-128k} & \textbf{8B-Instruct-128k}  \\
        \midrule
        
        16,384     & 34.39 & 83.64 & 77.92 & \textbf{86.28} & 91.5 \\
        32,768     & 32.90 & 79.21 & 72.71 & \textbf{82.27} & 89.8 \\
        65,536     & 35.01 & 63.80 & 66.42 & \textbf{75.95} & 87.6 \\
        131,072    & 20.07 & 23.61 & 59.26 & \textbf{63.57} & 81.7 \\
        \bottomrule
    \end{tabular}
    }
    \caption{Average RULER benchmark scores up to 128k context length for aligned Nemotron-H 4B and other instruction-tuned models in a similar size range.}
    \label{tab:long_context_results}
\end{table*}



\subsection{Evaluation Summary}

\begin{table}[h]
\centering
\resizebox{0.5\textwidth}{!}{
\begin{tabular}{l|c|c}
\hline
\textbf{Model} & \textbf{Garak Score} & \textbf{AEGIS Score} \\
\hline
Nemotron-H-8B & 70.75 & 99.83 \\
\hline
Nemotron-H-4B & 67.77 & 98.17 \\
\hline
\end{tabular}
}
\caption{Safety scores before and after compression.}
\label{tab:safety_scores}
\end{table}

\begin{figure}[ht]
    \includegraphics[width=\columnwidth]{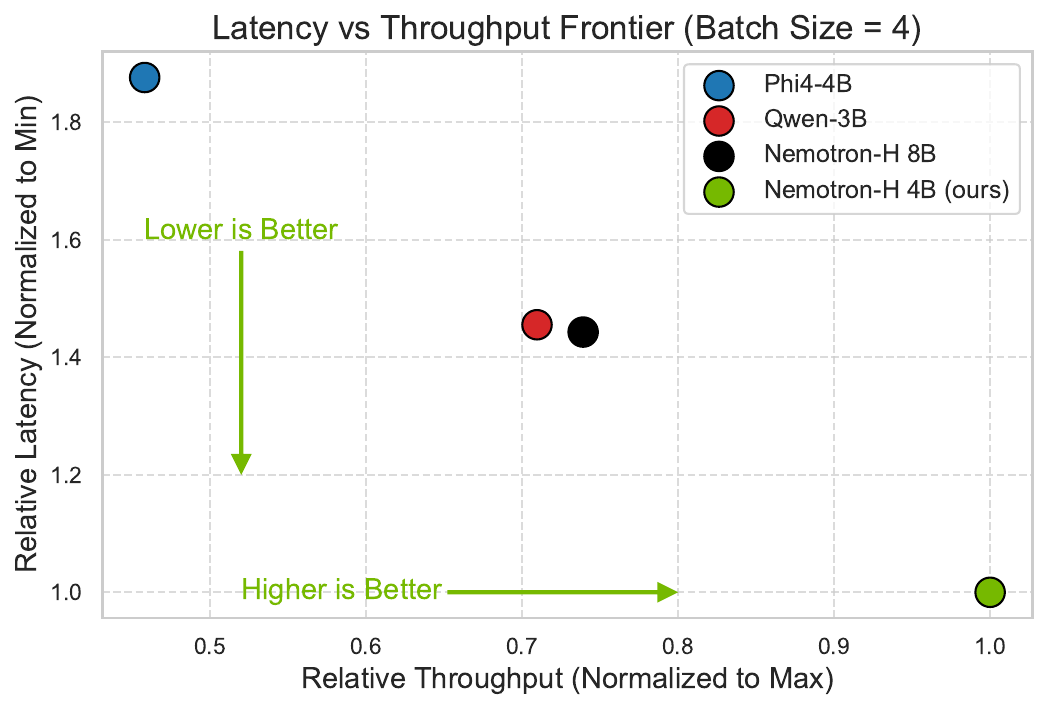}
    \caption{Throughput and latency comparisons across four models: Phi-4-Mini-4B, Qwen-2.5-3B, Nemotron-H 8B, and Nemotron-H 4B (ours). Relative throughput and latency represents are measured for an input and output context length of 65536 and 1024, respectively.}
    \label{fig:latency_tput}
\end{figure}

\noindent\textbf{Evaluation Summary.}
Tables~\ref{tab:base_model_results} to~\ref{tab:safety_scores} present accuracy comparisons between our compressed 4B hybrid model, other similar-sized community models, and the parent 8B hybrid model. 
As shown in Tables, our 4B model retains over 96\% of the original 8B model’s accuracy, including safety scores on Garak and AEGIS, while improving throughput by $\sim$1.4x. Compared to other similarly sized community models, it delivers state-of-the-art accuracy across knowledge, math, coding, commonsense reasoning, and reading comprehension tasks, despite being trained on up to $\sim$40x fewer tokens. It also achieves $\sim$2.2x higher throughput and $\sim$1.8x lower latency than the second-best Phi-4-4B model (Figures~\ref{fig:teaser} and~\ref{fig:latency_tput}). The aligned version further leads in math, coding, instruction following, and tool-use tasks.

To assess long-context capabilities, we use the RULER benchmark~\citep{hsieh2024rulerwhatsrealcontext}. As shown in Table~\ref{tab:long_context_results}, our model demonstrates strong performance and achieves the highest scores at context lengths up to 128k tokens

Figure~\ref{fig:latency_tput} compares latency and throughput across four models: Phi-4-Mini-4B, Qwen-2.5-3B, Nemotron-H 8B, and \textbf{Nemotron-H 4B (ours)}. 
Our model achieves the best performance on both axes—delivering the fastest time-to-first-token and highest throughput—effectively advancing the latency-throughput Pareto frontier. 

In summary, our compression approach successfully produces a model with state-of-the-art accuracy while significantly improving inference speed and reducing training costs. 



\subsection{Generalizability to Mamba2}

To evaluate the generalizability of our compression strategy to other models, we apply it to the Mamba2 1.3B model \cite{dao2024transformers}. We prune the model to 780M parameters via SSM and embedding pruning, and then subsequently train the pruned model on 10.5B tokens. We compare our pruned 780M model to the Mamba2 780M and 1.3B models trained from scratch on 300B tokens \cite{dao2024transformers}. 

As shown in Table \ref{tab:mamba2_results}, our compressed 780M model, despite being trained on significantly fewer tokens (10.5B vs. 300B), outperforms the 780M model trained from scratch and achieves an average score comparable to the original 1.3B model. These results provide further insights into the generalizability of our compression method.

\begin{table*}[h!]
\centering
\small 
\resizebox{0.75\textwidth}{!}{\begin{tabular}{lrrr}
\toprule
\textbf{Benchmark} & \textbf{Mamba2 780M} & \textbf{Mamba2 1.3B} & \textbf{Compressed 780M} \\
\midrule
ARC Challenge     & 28.6                 & 33.2                 & 34.2                     \\
ARC Easy          & 54.7                 & 60.6                 & 60.4                     \\
CommonsenseQA    & 19.6                 & 20.9                 & 26.4                     \\
HellaSwag          & 54.7                 & 59.9                 & 50.4                     \\
OpenbookQA         & 36.4                 & 37.0                 & 34.2                     \\
PIQA               & 72.1                 & 73.5                 & 71.3                     \\
RACE               & 21.8                 & 24.8                 & 32.5                     \\
Social IQA        & 41.0                 & 42.9                 & 41.1                     \\
TruthfulQA MC2    & 38.1                 & 36.1                 & 38.9                     \\
Winogrande         & 58.0                 & 60.1                 & 57.7                     \\
\midrule
\textbf{Avg}       & \textbf{42.5}        & \textbf{44.9}        & \textbf{44.7}            \\
\midrule
Tokens             & 300B                 & 300B                 & 10.5B                    \\
\bottomrule
\end{tabular}}
\caption{Comparison of our compressed Mamba2 780M model against Mamba2 780M and 1.3B models trained from scratch \cite{dao2024transformers}. Despite being trained on significantly fewer tokens (10.5B vs. 300B), our compressed model achieves a better average score than the 780M baseline.}
\label{tab:mamba2_results}
\end{table*}




%% file: main.bbl
\begin{thebibliography}{10}

\bibitem{vaswani2017attention}
Ashish Vaswani, Noam Shazeer, Niki Parmar, Jakob Uszkoreit, Llion Jones, Aidan~N Gomez, {\L}ukasz Kaiser, and Illia Polosukhin.
\newblock Attention is all you need.
\newblock {\em Advances in neural information processing systems}, 30, 2017.

\bibitem{gu2023mamba}
Albert Gu and Tri Dao.
\newblock Mamba: Linear-time sequence modeling with selective state spaces.
\newblock {\em arXiv preprint arXiv:2312.00752}, 2023.

\bibitem{dao2024transformers}
Tri Dao and Albert Gu.
\newblock Transformers are ssms: Generalized models and efficient algorithms through structured state space duality.
\newblock {\em arXiv preprint arXiv:2405.21060}, 2024.

\bibitem{hinton2015distilling}
Geoffrey Hinton, Oriol Vinyals, and Jeff Dean.
\newblock {Distilling the Knowledge in a Neural Network}.
\newblock {\em arXiv preprint arXiv:1503.02531}, 2015.

\bibitem{Minitron}
Saurav Muralidharan, Sharath~Turuvekere Sreenivas, Raviraj Joshi, Marcin Chochowski, Mostofa Patwary, Mohammad Shoeybi, Bryan Catanzaro, Jan Kautz, and Pavlo Molchanov.
\newblock Compact language models via pruning and knowledge distillation.
\newblock {\em arXiv preprint arXiv:2407.14679}, 2024.

\bibitem{bercovich2024puzzledistillationbasednasinferenceoptimized}
Akhiad Bercovich, Tomer Ronen, Talor Abramovich, Nir Ailon, Nave Assaf, Mohammad Dabbah, Ido Galil, Amnon Geifman, Yonatan Geifman, Izhak Golan, Netanel Haber, Ehud Karpas, Roi Koren, Itay Levy, Pavlo Molchanov, Shahar Mor, Zach Moshe, Najeeb Nabwani, Omri Puny, Ran Rubin, Itamar Schen, Ido Shahaf, Oren Tropp, Omer~Ullman Argov, Ran Zilberstein, and Ran El-Yaniv.
\newblock {Puzzle: Distillation-Based NAS for Inference-Optimized LLMs}, 2024.

\bibitem{tang2025darwinlm}
Shengkun Tang, Oliver Sieberling, Eldar Kurtic, Zhiqiang Shen, and Dan Alistarh.
\newblock Darwinlm: Evolutionary structured pruning of large language models.
\newblock {\em arXiv preprint arXiv:2502.07780}, 2025.

\bibitem{munoz2025mamba}
J~Pablo Mu{\~n}oz, Jinjie Yuan, and Nilesh Jain.
\newblock Mamba-shedder: Post-transformer compression for efficient selective structured state space models.
\newblock {\em arXiv preprint arXiv:2501.17088}, 2025.

\bibitem{ghattas2025pruning}
Tamer Ghattas, Michael Hassid, and Roy Schwartz.
\newblock On pruning state-space llms.
\newblock {\em arXiv preprint arXiv:2502.18886}, 2025.

\bibitem{blakeman2025nemotron}
Aaron Blakeman, Aarti Basant, Abhinav Khattar, Adithya Renduchintala, Akhiad Bercovich, Aleksander Ficek, Alexis Bjorlin, Ali Taghibakhshi, Amala~Sanjay Deshmukh, Ameya~Sunil Mahabaleshwarkar, et~al.
\newblock Nemotron-h: A family of accurate and efficient hybrid mamba-transformer models.
\newblock {\em arXiv preprint arXiv:2504.03624}, 2025.

\bibitem{jamba2024hybrid}
Opher Lieber, Barak Lenz, Hofit Bata, Gal Cohen, Jhonathan Osin, Itay Dalmedigos, et~al.
\newblock Jamba: A hybrid transformer-mamba language model.
\newblock {\em arXiv preprint arXiv:2403.19887}, 2024.

\bibitem{glorioso2024zamba}
Paolo Glorioso, Quentin Anthony, and Yury Tokpanov.
\newblock Zamba: A compact 7b ssm hybrid model.
\newblock {\em arxiv preprint arXiv:2405.16712}, 2024.

\bibitem{wang2024model}
Wenxiao Wang, Wei Chen, Yicong Luo, Yongliu Long, Zhengkai Lin, Liye Zhang, Binbin Lin, Deng Cai, and Xiaofei He.
\newblock Model compression and efficient inference for large language models: A survey.
\newblock {\em arXiv preprint arXiv:2402.09748}, 2024.

\bibitem{hoefler:2021}
Torsten Hoefler, Dan Alistarh, Tal Ben-Nun, Nikoli Dryden, and Alexandra Peste.
\newblock Sparsity in {D}eep {L}earning: {P}runing and growth for efficient inference and training in neural networks.
\newblock {\em arXiv preprint arXiv:2102.00554}, 2021.

\bibitem{luo:2017}
Jian-Hao Luo, Jianxin Wu, and Weiyao Lin.
\newblock Thinet: A filter level pruning method for deep neural network compression.
\newblock In {\em Proceedings of the IEEE international conference on computer vision}, pages 5058--5066, 2017.

\bibitem{he:2018}
Yang He, Guoliang Kang, Xuanyi Dong, Yanwei Fu, and Yi~Yang.
\newblock Soft filter pruning for accelerating deep convolutional neural networks.
\newblock {\em arXiv preprint arXiv:1808.06866}, 2018.

\bibitem{xia2023sheared}
Mengzhou Xia, Tianyu Gao, Zhiyuan Zeng, and Danqi Chen.
\newblock Sheared llama: Accelerating language model pre-training via structured pruning.
\newblock In {\em The Twelfth International Conference on Learning Representations}, 2023.

\bibitem{ashkboos2023slicegpt}
Saleh Ashkboos, Maximilian~L Croci, Marcelo~Gennari do~Nascimento, Torsten Hoefler, and James Hensman.
\newblock Slicegpt: Compress large language models by deleting rows and columns.
\newblock In {\em The Twelfth International Conference on Learning Representations}, 2023.

\bibitem{men2024shortgpt}
Xin Men, Mingyu Xu, Qingyu Zhang, Bingning Wang, Hongyu Lin, Yaojie Lu, Xianpei Han, and Weipeng Chen.
\newblock {ShortGPT: Layers in Large Language Models are More Redundant Than You Expect}, 2024.

\bibitem{yang2024laco}
Yifei Yang, Zouying Cao, and Hai Zhao.
\newblock Laco: Large language model pruning via layer collapse.
\newblock {\em arXiv preprint arXiv:2402.11187}, 2024.

\bibitem{kim2024shortened}
Bo-Kyeong Kim, Geonmin Kim, Tae-Ho Kim, Thibault Castells, Shinkook Choi, Junho Shin, and Hyoung-Kyu Song.
\newblock Shortened {LL}a{MA}: A simple depth pruning for large language models.
\newblock In {\em ICLR 2024 Workshop on Mathematical and Empirical Understanding of Foundation Models}, 2024.

\bibitem{flap}
Yongqi An, Xu~Zhao, Tao Yu, Ming Tang, and Jinqiao Wang.
\newblock Fluctuation-based adaptive structured pruning for large language models.
\newblock In {\em Proceedings of the AAAI Conference on Artificial Intelligence}, volume~38, pages 10865--10873, 2024.

\bibitem{sreenivas2024llmpruningdistillationpractice}
Sharath~Turuvekere Sreenivas, Saurav Muralidharan, Raviraj Joshi, Marcin Chochowski, Ameya~Sunil Mahabaleshwarkar, Gerald Shen, Jiaqi Zeng, Zijia Chen, Yoshi Suhara, Shizhe Diao, Chenhan Yu, Wei-Chun Chen, Hayley Ross, Oluwatobi Olabiyi, Ashwath Aithal, Oleksii Kuchaiev, Daniel Korzekwa, Pavlo Molchanov, Mostofa Patwary, Mohammad Shoeybi, Jan Kautz, and Bryan Catanzaro.
\newblock {LLM Pruning and Distillation in Practice: The Minitron Approach}, 2024.

\bibitem{nvidia2024nemotron4340btechnicalreport}
Bo~Adler, Niket Agarwal, Ashwath Aithal, Dong~H Anh, Pallab Bhattacharya, Annika Brundyn, Jared Casper, Bryan Catanzaro, Sharon Clay, Jonathan Cohen, et~al.
\newblock Nemotron-4 340b technical report.
\newblock {\em arXiv preprint arXiv:2406.11704}, 2024.

\bibitem{shen2024nemoalignerscalabletoolkitefficient}
Gerald Shen, Zhilin Wang, Olivier Delalleau, Jiaqi Zeng, Yi~Dong, Daniel Egert, Shengyang Sun, Jimmy Zhang, Sahil Jain, Ali Taghibakhshi, Markel~Sanz Ausin, Ashwath Aithal, and Oleksii Kuchaiev.
\newblock Nemo-aligner: Scalable toolkit for efficient model alignment, 2024.

\bibitem{hsieh2024rulerwhatsrealcontext}
Cheng-Ping Hsieh, Simeng Sun, Samuel Kriman, Shantanu Acharya, Dima Rekesh, Fei Jia, Yang Zhang, and Boris Ginsburg.
\newblock Ruler: What's the real context size of your long-context language models?, 2024.

\end{thebibliography}
